\documentclass[tablecaption=bottom,wcp]{jmlr} 

\usepackage{booktabs}
\usepackage[load-configurations=version-1]{siunitx} 

\theorembodyfont{\upshape}
\theoremheaderfont{\scshape}
\theorempostheader{:}
\theoremsep{\newline}

\jmlrproceedings{AABI 2021}{4th Symposium on Advances in Approximate Bayesian Inference, 2021}

\title[PAC-Bayesian matrix completion]{PAC-Bayesian Matrix Completion with a Spectral Scaled Student Prior}

\author{\Name{The Tien Mai}
	\Email{the.t.mai@ntnu.no}\\
	\addr Department of Mathematical Sciences, \\
	Faculty of Information Technology and Electrical Engineering, \\
	Norwegian University of Science and Technology, Trondheim, Norway.}

\usepackage{algorithm,algorithmicx,algpseudocode}
\usepackage{hyperref}
\hypersetup{  colorlinks=true,
	linkcolor=blue,
	filecolor=magenta,      
	urlcolor=cyan,
	citecolor=blue}
\newtheorem{assume}{Assumption}

\begin{document}
	
	\maketitle
	
	\begin{abstract}
		We study the problem of matrix completion in this paper.  A spectral scaled Student prior is exploited to favour the underlying low-rank structure of the data matrix.  We provide a thorough theoretical investigation for our approach through PAC-Bayesian bounds. More precisely, our PAC-Bayesian approach enjoys a minimax-optimal oracle inequality which guarantees that our method works well under model misspecification and under general sampling distribution. Interestingly,  we also provide efficient gradient-based sampling implementations for our approach by using Langevin Monte Carlo. More specifically, we show that our algorithms are significantly faster than Gibbs sampler in this problem.  To illustrate the attractive features of our inference strategy, some numerical simulations are conducted and an application to image inpainting is demonstrated.
	\end{abstract}
	
	
	\section{Introduction}
	Arising in various applications such as recommender systems \cite{bennett2007netflix,xiong2010temporal,adomavicius2011context}, genotype imputation \cite{chi2013genotype,jiang2016sparrec},  image processing \cite{cabral2014matrix,luo2015multiview,he2014image} and quantum state tomography \cite{gross2010quantum,mai2017pseudo}, matrix completion aims at rebuilding a matrix from its partially observed entries. Most of the recent methods for this problem are usually based on penalized optimizations which are considered both from theoretical and computational point of views.  The seminal results were done in \cite{candes2009exact,candes2010power,candes2010matrix, koltchinskii2011nuclear,negahban2012restricted}.  Several efficient algorithms had also been proposed and studied as in~\cite{mazumder2010spectral,recht2013parallel,hastie2015matrix}. 
	
	On the other hand,  various Bayesian approaches have also been proposed in the problem of matrix completion largely from a computational direction~\cite
	{lim2007variational, salakhutdinov2008bayesian,
		zhou2010nonparametric,alquier2014bayesian,
		lawrence2009non,cottet20181,babacan2012sparse,yang2018fast}.  These Bayesian estimators are mostly based on conjugate low-rank factorization priors which allow to use Gibbs sampling~\cite{alquier2014bayesian, salakhutdinov2008bayesian}.  Nevertheless, these Gibbs samplers require to calculate a number of matrix inversions or singular value decompositions at each iteration which is costly and can slow down significantly the algorithm for large data.  Variational Bayes methods based on optimization have been also considered in this problem~\cite{lim2007variational,babacan2012sparse,yang2018fast}. 
	
	However, the theoretical understanding of Bayesian estimators has received quite limited attention, up to ourknowledge,  \cite{mai2015} and \cite{alquier2020concentration} are the only prominent works. More specifically,  they showed that a Bayesian estimator with a low-rank factorization prior reaches the minimax-optimal rate up to a logarithmic factor.  The paper \cite{alquier2020concentration} further shows that the same rate can be obtained by using a Variational Bayesian estimator and the concentration rate of posterior is also studied in their works.

	In this paper, we study the problem of Bayesian matrix completion where a spectral scaled Student is exploited to favour the (approximate) low-rank structure of the underlying matrix. We prove that our PAC-Bayesian estimator enjoys a general minimax-optimal oracle inequality.  
	As a result,  it shows that our estimator works well in general cases which are under model misspecification and under general sampling distributon. While this theory is similar to \cite{mai2015}, where a different prior (a low-rank factorizaton prior) is studied,  our result presents an absolute improvement over that paper. More specifically,  the leading constant in our minimax-optimal oracle inequality is strictly smaller than 2 while the leading constant in \cite{mai2015} is 3.  Up to our knowledge, a sharp minimax-optimal oracle inequality with leading constant 1 has not yet been obtained for Bayesian matrix completion.

	\section{Bayesian matrix completion}
	\label{sc_model_bmc}
	\subsection{Model}
	Let $ M^* \in  \mathbb{R}^{m\times p} $ be an unknown  (expected to be low-rank) matrix of interest.  We observe a random subset of noisy entries of $M^*$ as 
	\begin{equation}
	\label{main model}
	Y_{ij} = M^*_{ij} + \mathcal{E}_{ij},   \quad (i,j) \in \Omega
	\end{equation}
	where $\Omega $ is a subset of indices $\lbrace1, \ldots, m \rbrace\times\lbrace1, \ldots,p \rbrace $ and $ \mathcal{E}_{ij} $ are independently generated noise at the location $(i,j) $ with $ \mathbb{E} (\mathcal{E}_{ij}) =0 $.  Let $ \Pi_{ij}$ denote the probability to observe the $ (i,j) $-th entry. Then, the problem of recovering $ M^* $ with $ n = |\Omega| < mp $ under  the assumption that $ {\rm rank}(M^*) \ll \min(m,p) $ is called the noisy matrix completion problem.
	
	
	We consider the following (pseudo-)Bayesian mean estimator with a given prior distribution $  \pi (dM) $,
	\begin{equation}
	\label{bmc_estimator}
	\hat{M}_{\lambda} = \int M \hat{\rho}_{\lambda}(d M),
	\end{equation}
	where 
	$
	\hat{\rho}_{\lambda}(d M)
	\propto   
	\exp (-\lambda r(M)) \pi (dM)
	$
	is the posterior in which 
	$
	r(M) = \sum_{ (i,j) \in \Omega } \left( Y_{ij} - M_{ij } \right)^2/n  .
	$

	The choice $ \lambda = n/(2\sigma^2) $ is corresponding exactly to the Bayesian mean estimator that would be obtained for a Gaussian noise. However,  by using $ \lambda $, this will allow us to obtain the optimality of the estimator under a wider class of noises. Moreover,  this kind of fractional posterior has been shown to work well in misspecification model \cite{alquier2020concentration,
		bhattacharya2019bayesian,
		grunwald2017inconsistency,
		bissiri2016general} which in our setup can be used for approximate low-rank model.

	\subsection{Low-rank promoting prior: a spectral scaled Student prior }
	We consider the following prior,
	\begin{align}  
	\label{prior_scaled_Student}
	\pi(M) 
	\propto 
	\det (\tau^2 \mathbf{I}_{m} + MM^\intercal )^{-(p+m+2)/2}
	\end{align}
	where $ \tau > 0 $ is a tuning parameter and $ \mathbf{I}_{m} $ is the $ m \times m $ identity matrix . 
	
	To illustrate that this prior has the potential to encourage the low-rankness of $ M $, one can check that 
	\begin{align*}
	\pi(M) 
	\propto \prod_{j=1}^{m}  (\tau^2  + s_j(M)^2 )^{- (p+m+2)/2 },
	\end{align*}
	where $ s_j(M) $ denotes the $j $-th largest singular value of $ M $. It is well known that the log-sum function $ \sum_{j=1}^m \log (\tau ^2  + s_j(M)^2) $ encourages a sparsity on $\{s_j(M)\}$, see \cite{candes2008enhancing,yang2018fast}.  Alternatively, one can recognize a scaled Student distribution evaluated at $ s_j(M)  $ in the last display above which induces sparsity on $\{s_j(M)\}$, \cite{dalalyan2012sparse}. Thus the resulting matrix $ M $ enjoys a low-rank structrure, approximately.  
	
	Although this prior is not conjugate in our problem, it is particularly convenient to implement the Langevin Monte Carlo algorithm, a gradient-based sampling method,  see Section \ref{sc_LMC}.

	\subsection{Theoretical guarantees}
	\label{sc_theory}
	
	Before we present the theoretical guarantees for our procedure, let us formulate some assumptions.
	
	\begin{assume}
		\label{bounded assume}
		There is a known $ L>0 $ such that
		$
		\| M^* \|_{\infty} = \sup\limits_{i,j} \vert M^*_{ij} \vert   \leq  L  < +\infty.
		$
	\end{assume}
	
	\begin{assume}
		\label{bruit-Pac}
		The noise variables  $ \mathcal{E}_1, \ldots, \mathcal{E}_n$ are independent 
		and independent of $X_1,\ldots,X_{n}$. There exist two known
		constants $\sigma>0$ and $\xi>0$ such that
		$$ \mathbb{E} (\mathcal{E}_{i}^{2})\leq \sigma^{2} $$
		$$ \forall k\geq 3,\quad \mathbb{E} (|\mathcal{E}_{i}|^{k}) \leq \sigma^{2} k! \xi^{k-2}.$$
	\end{assume}

	
	For any matrix $ A_{m\times p} $, let $\|A\|_F$ denote the Frobenius norm, i.e, $\|A\|_F^2
	= {\rm Tr}(A^T A)$.  We define a (general-Frobenius) ``norm'' as follow
	$
	\|A\|^2_{F,\Pi} = \sum_{ij} (A_{ij})^2 \Pi_{ij}$.
	Note that when the sampling distribution $ \Pi $ is uniform, then 
	$
	\|A\|^2_{F,\Pi}
	= 
	\|A\|^2_F/(mp). 
	$

	Put $ C_1 := 2 \left[ 4 \sigma^{2} + (3L)^2 \right] $ and $ C_2:=  12 L (2 \xi + 3 L ) $ and take $C$ such that $ C > C_2 + 3C_1/2 $.  We are now ready to state our main theoretical result and the proof of this theorem is postponed to Appendix \ref{sc_proofs}.

	\begin{theorem}
		\label{thrm_main}
		Let Assumption \ref{bounded assume} and \ref{bruit-Pac} be satisfied and take $ \lambda = \lambda^* := n/C $. Then, with probability at least $ 1 - \epsilon, \epsilon\in (0,1) $, one has for any matrix $ \bar{M} $ with its rank at most $r$ that
		\begin{equation}
		\label{bound_main}
		\| \hat{M}_{\lambda^*} - M^*  \|^2_{F,\Pi} 
		\leq
		\inf_{ \bar{M} } \,
		(1 + \delta) \| \bar{M} - M^*  \|^2_{F,\Pi} 
		+ 
		\frac{ \mathcal{C}  r(m+p) \log \left(mp \right)  } {n} 
		+
		\frac{\mathcal{C} } {n}\log\frac{2}{\varepsilon}
		,
		\end{equation}
		where $ \delta $ is in $ (0,1) $ and $ \mathcal{C} > 6C $  is a universal constant that depends on $ \sigma^2, L, \xi $ only.
	\end{theorem}

	We remark that our oracle inequality \eqref{bound_main} comes with a leading constant $ 1+\delta < 2 $ (thank to a more careful calculation) with $ 0< \delta < 1 $ that is absolutely smaller than the leading constant $ 3 $ in Theorem 1 in \cite{mai2015}. Thus our work presents an improved oracle inequality for Bayesian matrix completion.  We would like to note that a sharp oracle inequality with a leading constant 1 is not yet obtained for Bayesian matrix completion and remains as an  important open question.

	We note that Theorem \ref{thrm_main} is stated under a general setting. More specifically, it holds without any assumption on the sampling distribution of the observations as done in \cite{mai2015} while other works require some, see e.g. \cite{foygel2011learning, klopp2014noisy,
		negahban2012restricted}. Moreover,  Theorem \ref{thrm_main} can also be used in various setup where the underlying matrix $ M^* $ is for example low-rank, approximate low-rank\ldots Several special cases are derived in the Appendix \ref{sc_proofs}. 

	It is remarked that the convergence rate $ r(m+p) \log \left(mp \right) /n $ is minimax-optimal up to a log-factor.  A lower bound for low-rank matrix completion is provided in   \cite{koltchinskii2011nuclear} that is $ r(m+p)/n $,  whereas the sharp upper bounds are obtained by penalized minimization methods in \cite{klopp2015matrix,chen2019inference}. Up to our knowledge, however, a sharp rate for Bayesian estimators in the problem of matrix completion still remains open. The paper \cite{mai2021numerical}, based on numerical comparisons between Bayesian methods and a de-biased estimator in \cite{chen2019inference} which sharply reached the minimiax-optimal rate, conjectures that the Bayesian methods could actually reach this rate sharply and the additonal logarithmic factor could be due to the teachnical proofs.

	\section{Numerical Studies}
	\label{sc_LMC}
	\subsection{Unadjusted Langevin Monte Carlo algorithm}
	Let $  \mathcal{P}_\Omega (\cdot) : \mathbb{R}^{m\times p}  \mapsto  \mathbb{R}^{m\times p}  $ be the orthogonal projection onto the observed entries in the index set $\Omega $ that $ \mathcal{P}_\Omega (Y)_{ij} = Y_{ij},   \text{ if } (i,j) \in \Omega, $ and 0 otherwise.

	We propose to compute an approximation of the posterior in \eqref{bmc_estimator} by a suitable version of the Langevin Monte Carlo algorithm, a gradient-based sampling method. 
	
	Let us remind that we have
	$$
	\nabla  \log   \hat{\rho}_{\lambda}( M)
	= 
	- \frac{2\lambda}{n} \mathcal{P}_\Omega (Y-M) 
	- (p+m+2) (\tau^2 \mathbf{I}_{m} + MM^\intercal )^{-1} M .
	$$
	In this work, we aim at using the constant step-size unadjusted Langevin Monte Carlo algorithm (denoted by LMC), see \cite{durmus2019high} for detail.  This algorithm is defined by selecting an initial matrix $M_0$ and then by using the recursion
	\begin{align}
	\label{langevinMC}
	M_{k+1} = M_{k} - h\nabla \log    \hat{\rho}_{\lambda}(M_k) +\sqrt{2h}\,W_k,\qquad
	k=0,1,\ldots,
	\end{align}
	where $h>0$ is the step-size and $ W_0, W_1,\ldots$ are independent random matrices with i.i.d. standard Gaussian entries.


	For small values of the step-size $ h $, the posterior mean $ \hat{M} = \sum_k^T M_k /T $   is very close to the integral \eqref{bmc_estimator} of interest. However, for some $h$ that may not be small enough, the Markov process can be transient and as a consequence the sum explodes \cite{roberts2002langevin}.  Several strategies are available to address this issue: one can take a smaller h and restart the algorithm or a Metropolis–Hastings correction can be included in the algorithm (denoted MALA, see details in the Appendix \ref{sc_proofs}). The Metropolis–Hastings approach ensures the convergence to the desired distribution, however,  the algorithm is greatly slowed down because of an additional acception/rejection step at each iteration. Taking a smaller $h $ also slows down the algorithm but we keep some control on its time of execution.

	\subsection{Simulation studies}
	\label{sc_numer}
	In order to access the behaviour of our algorithms, we first conducted
	a series of experiments with simulated data:
	\begin{itemize}
		\item Setting I: In the first setting,  a rank-$r $ matrix $ M^* $ is generated as the product of two rank-$r $ matrices, 
		$$ 
		M^* = U^*_{m\times r} (V^*_{p\times r})^\top ,
		$$
		where the entries of $U^* $ and $ V^* $ are i.i.d $ \mathcal{N} (0 , 1) $. With a missing rate $ \upsilon = 20\%, 50\% $ and $80\%$,  we observe the entries of the matrix $ M^* $ using a uniform sampling.  Then, this sampled set is 	corrupted by noise as in~\eqref{main model}, where $ \mathcal{E}_i $ are i.i.d $ \mathcal{N} (0 , 1) $.  The dimensions are alternated by fixing $ m= 100 $ and varying $ p = 100 $ and 500.  The rank $r$ is varied between $r = 2 $ and $r = 5 $.
		\item Setting II: The second series of simulations is similar to the first one, except that the matrix $M^*$ is no longer rank-$r$, but it can be well approximated by a rank-$r$ matrix:
		\[
		M^*=U^*_{m\times r} (V^*_{m\times r})^\top + \frac{1}{10}
		(A_{m\times50})(B_{m\times50})^\top
		\]
		where the entries of $A $ and $B$ are i.i.d $ \mathcal{N} (0 , 1) $.
	\end{itemize}
	
	We compare our method with the state-of-the-art method in Bayesian matrix completion from \cite{alquier2014bayesian,alquier2020concentration} which employed a Gibbs sampler algorithm. This method is denoted by `Gibbs'.
	
	For each setting, we simulate 50 data sets (simulation replicates). Then, we report the average and the standard deviation for a measure of error of each method over the replicates.  The performance of a method (say $\widehat{M}$) is measured by the mean squared error (MSE) per entry
	$
	{\rm MSE} :=  \|\widehat{M} - M^*\|_F^2/(mp)
	$
	and the normalized mean square error (NMSE)
	$
	{\rm NMSE} :=  \|\widehat{M} - M^*\|_F^2 / \| M^*\|_F^2 .
	$
	We also measure the prediction error by using
	$
	{\rm Pred} :=  \|   \mathcal{P}_{\bar{\Omega} } ( \widehat{M} - M^*)  \|_F^2/(mp- n),
	$
	where $ \bar{\Omega} $ is the set of un-observed entries.
	
	The results are reported in Table \ref{tb_setting1} and \ref{tb_setting2}. The choice of the step-size parameters is set as $ h = 1/(pm400) $ which is selected such that the acceptance rate of MALA is approximate 0.5. We fixed $ \tau = 1 $ in all settings.  The LMC, MALA and Gibbs sampler are used with tuning parameter $\lambda = 1/(4\sigma^2) $.  In this simulation, LMC and MALA are initialized by using the output from the Gibbs sampler (after 50 iterations run). The parameters for the prior of the `Gibbs' method are $K=10,a=1, b=1/100$.  These algorithms are run with $ T = 200 $ iterations and we take the first 100 steps as burn-in.

\begin{figure}[!h]
		\centering
		\includegraphics[scale=.6]{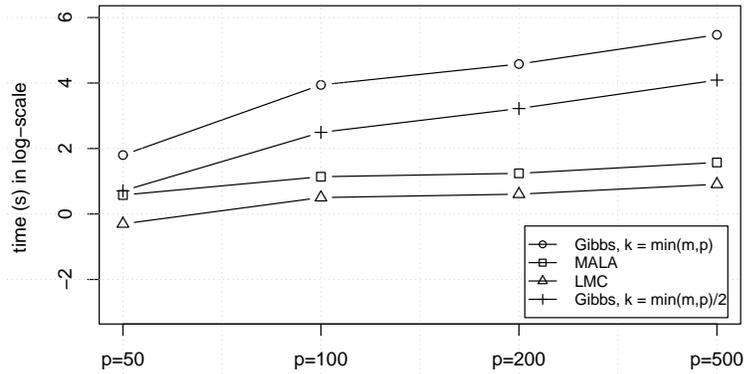}
		\caption{Plot to compare the running times for 20 iterations of LMC, MALA and Gibbs sampler with fixed $m =100, r = 2$ and  20\% of the entries are missing.}
		\label{fg_runtime}
	\end{figure}

	Resuts from our simulations in Table \ref{tb_setting1} and \ref{tb_setting2} show that our proposed algortihms (LMC and MALA) perform quite similar to the Gibbs sampler. However,  it is shown that LMC and MALA are much more faster than the Gibbs sampler, see Figure \ref{fg_runtime}. More specifically, we compare the running time of the LMC, MALA against the Gibbs sampler with $ k =\min(m,p) $ and $ k =\min(m,p)/2 $, in which we fixed $ m=100, r =2,   \upsilon = 20\%$ and $p$ is varied by $50, 100, 200, 500 $.

	\subsection{Real application to Image inpainting}
	\label{sc_lena}
	We further applied our proposed algorithms on image inpainting to evaluate their performance on real data. The aim of image inpainting is to complete an image with missing pixels.  Here, we applied our methods on the well-known benchmark image Lena \cite{gonzalez2007digital} which is of size $ 256\times 256 $.  This data has been used before in the context of matrix completion in \cite{yang2018fast}. The LMC and MALA are initialized from an output from the SoftImpute method in \cite{mazumder2010spectral} which is available in \cite{pkg_filling,pkg_softimpute}.  
	
	We consider the cases where $  \upsilon =20\% $ and $ \upsilon = 50\% $ of pixels are missing uniformly at random.  The experiments are repeated 30 times and the average of the considered errors is reported. The original Lena $256\times 256$ image with missing pixels and these images recovered by respective algorithms and the estimation errors results are given in Table \ref{tb_lena}.  
	
\begin{figure}[!h]
		\centerline{\includegraphics[scale=.6]{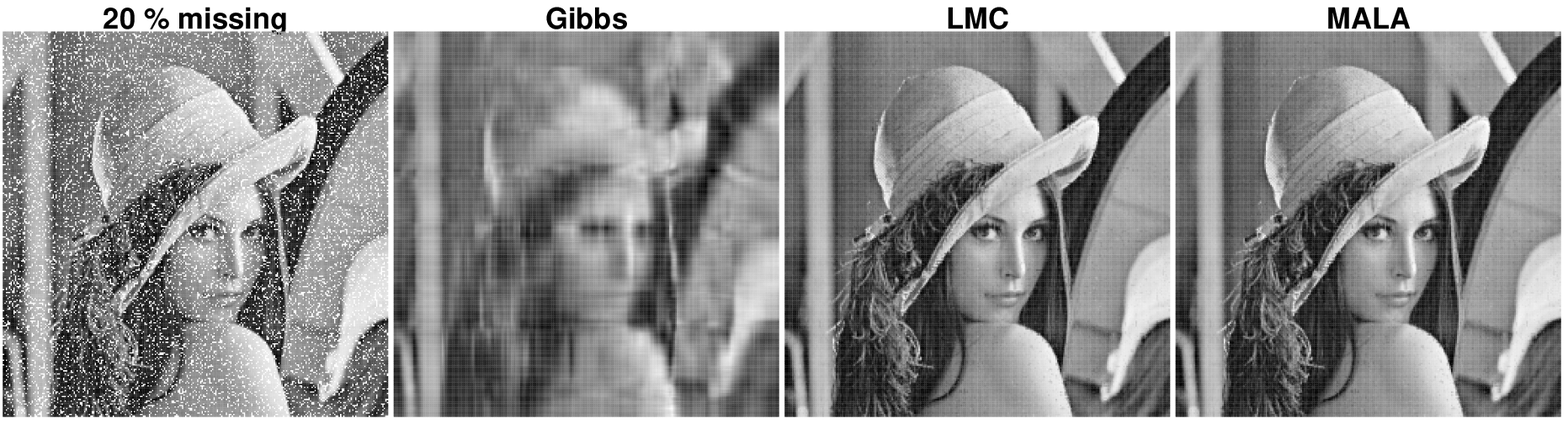} } 
		\centerline{\includegraphics[scale=.6]{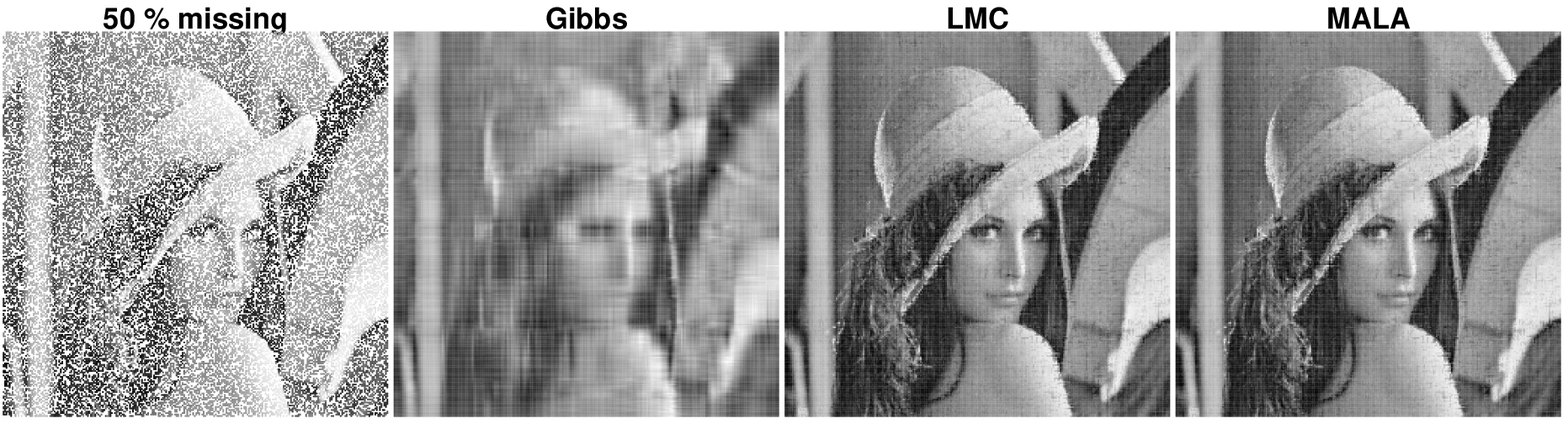} }
		\caption{The original Lena $256\times 256$ image with missing pixels and these images recovered by respective algorithms.}
		\label{fg_lena}
	\end{figure}

	From the results in Table \ref{tb_lena} and Figure \ref{fg_lena},  we see that our proposed methods outperform the method based on Gibbs sampler. This can be explained as that the Gibbs sampler algorithm based on Gaussian low-rank factorizaton prior where all the entries of the image's matrix are positive and thus the Gaussian low-rank factorizaton prior may not be an appropriate prior for image inpainting.  On the other hand, our approach indicates that the spectral scaled student prior can capture the (apprioximate) low-rank structure and the smoothness of the image data matrix and as a consequence it helps obtain a better visual quality image, see Figure \ref{fg_lena}.

	\section{Conclusion}
	\label{sc_conclusion}
	
	In this paper, we have studied the problem of Bayesian matrix completion by using a spectral scaled Student prior to promoting the low-rank structure of the underlying matrix. We have provided a thorough theoretical evaluation for our Bayesian estimator under both model misspecification and general sampling distribution.  We have also provided efficient gradient-based sampling algorithms for our estimator by using Langevin Monte Carlo approach.  These attractive features of our inference strategy are demonstrated through numerical simulations and real application to image inpainting.


	\section*{Acknowledgements}
	 TTM is supported by the Norwegian Research Council grant number 309960 through the Centre for Geophysical Forecasting at NTNU. I would like to warmly thank the anonymous referees for their useful comments on this work.


	\begin{table}[!h]
		\footnotesize
		\caption{Simulation results for Setting I (exact low-rank). The mean and the standard deviation (in parentheses) of each error between the simulation replicates are presented. (MSE: average of mean square error; NMSE: average of normalized mean square error; Pred: average of prediction error.)}
		\centering
		\begin{tabular}{p{17mm}|ccc||ccc} 
			\hline\hline
			& \multicolumn{3}{ c | |}{$r = 2, p = 100,  \upsilon = 20\% $} 
			& \multicolumn{3}{ c  }{$r = 5, p = 100,  \upsilon = 20\% $} 
			\\
			Errors   & Gibbs & LMC & MALA  
			& Gibbs & LMC & MALA 
			\\ \hline
			$10^2\times$MSE & 5.220 (.355) & 5.223 (.355) & 5.222 (.355) 
			& 13.101 (.580) & 13.103 (.580) &  13.103 (.580) 
			\\ 
			$10^2\times$NMSE  & 2.715 (.470) & 2.717 (.470)  & 2.716 (.470)  
			& 2.644 (.309) & 2.644 (.309) & 2.644 (.309) 
			\\ 
			$10^2\times$Pred  & 5.548 (.498) & 5.550 (.498)  &  5.550 (.498) 
			& 14.934 (.890) & 14.936 (.890) &  14.935 (.890) 
			\\
			\hline
			& \multicolumn{3}{ c ||}{  $r = 2, p = 100,  \upsilon = 50\% $  } 
			& \multicolumn{3}{ c  }{$r = 5, p = 100,  \upsilon = 50\% $} 
			\\  
			& LMC & MALA & Gibbs 
			& LMC & MALA & Gibbs
			\\ \hline
			$10^2\times$MSE& 8.821 (.629) & 8.822 (.629) & 8.822 (.629) 
			& 23.890 (1.158) & 23.893 (1.159) &  23.891 (1.158) 
			\\
			$10^2\times$NMSE	& 4.496 (.766) & 4.497 (.766) & 4.496 (.766) 
			& 4.816 (.520) & 4.815 (.520) & 4.815 (.520) 
			\\
			$10^2\times$Pred  & 9.403 (.758) & 9.406 (.758) & 9.405 (.758) 
			& 27.222 (1.560) & 27.225 (1.562) &  27.223 (1.559) 
			\\
			\hline
			& \multicolumn{3}{ c| |}{  $r = 2, p = 100,  \upsilon = 80\% $  }
			& \multicolumn{3}{ c  }{ $r = 5, p = 100,  \upsilon = 80\% $}  
			\\  
			& LMC & MALA & Gibbs 
			& LMC & MALA & Gibbs
			\\ \hline
			$10^1\times$MSE & 2.954 (.295) & 2.955 (.295) & 2.954 (.295) 
			& 15.615 (2.593) & 15.616 (2.593) &  15.616 (2.593)
			\\
			$10^1\times$NMSE	& 1.538 (.288) & 1.538 (.288) & 1.538 (.288) 
			& 3.081 (.581) & 3.081 (.581)  & 3.081 (.581) 
			\\
			$10^1\times$Pred  & 3.146 (.340) & 3.146 (.340) & 3.146 (.340) 
			& 17.685 (2.999) & 17.685 (2.999) & 17.685 (2.999) 
			\\ 
			\hline\hline
			& \multicolumn{3}{ c || }{$r = 2, p = 500,  \upsilon = 20\% $} 
			& \multicolumn{3}{ c  }{$r = 5, p = 500,  \upsilon = 20\% $} 
			\\
			& LMC & MALA & Gibbs 
			& LMC & MALA & Gibbs
			\\ \hline
			$10^2\times$MSE & 3.092 (.132) & 3.092 (.132) & 3.092 (.132)
			& 7.763 (.202) & 7.764 (.202) & 7.763 (.202) 
			\\ 
			$10^2\times$NMSE  & 1.554 (.189) & 1.554 (.189) & 1.554 (.189) 
			& 1.554 (.101) & 1.554 (.101) & 1.554 (.101) 
			\\ 
			$10^2\times$Pred  & 3.222 (.153) & 3.223 (.153) & 3.222 (.153) 
			& 8.445 (.260) & 8.446 (.260) & 8.446 (.260) 
			\\
			\hline
			& \multicolumn{3}{ c ||}{  $r = 2, p = 500,  \upsilon = 50\% $  } 
			& \multicolumn{3}{ c  }{$r = 5, p = 500,  \upsilon = 50\% $} 
			\\  
			& LMC & MALA & Gibbs 
			& LMC & MALA & Gibbs
			\\ \hline
			$10^2\times$MSE	& 5.131 (.234) & 5.131 (.234) & 5.131 (.234) 
			& 13.270 (.393) & 13.270 (.393) & 13.270 (.393) 
			\\
			$10^2\times$NMSE	& 2.657 (.298) & 2.657 (.298) & 2.657 (.298) 
			& 2.648 (.208) & 2.648 (.208) & 2.648 (.208)
			\\
			$10^2\times$Pred & 5.369 (.261) & 5.369 (.261) & 5.369 (.261) 
			& 14.411 (.472) & 14.411 (.472) & 14.411 (.472) 
			\\
			\hline
			& \multicolumn{3}{ c| |}{  $r = 2, p = 500,  \upsilon = 80\% $  }
			& \multicolumn{3}{ c  }{ $r = 5, p = 500,  \upsilon = 80\% $}  
			\\  
			& LMC & MALA & Gibbs 
			& LMC & MALA & Gibbs
			\\ \hline
			$10^1\times$MSE	& 1.512 (.065) & 1.512 (.065) & 1.512 (.065) 
			& 4.6974 (.1718) & 4.6975 (.1718)  & 4.6974 (.1718) 
			\\
			$10^1\times$NMSE	& 0.766 (.077) & 0.766 (.077) & 0.766 (.077) 
			& .9416 (.0756) & .9417 (.0756) & .9417 (.0756) 
			\\
			$10^1\times$Pred  & 1.582 (.074) & 1.582 (.074)  & 1.582 (.074) 
			& 5.0958 (.2001) & 5.0959 (.2001) & 5.0959 (.2001) 
			\\ \hline\hline
		\end{tabular}
		\label{tb_setting1}
	\end{table}

	\begin{table}[!h]
		\footnotesize
		\caption{Simulation results for Setting II (approximate low-rank). The mean and the standard deviation (in parentheses) of each error between the simulation replicates are presented. (MSE: average of mean square error; NMSE: average of normalized mean square error; Pred: average of prediction error.)}
		\centering
		\begin{tabular}{p{17mm}|ccc||ccc} 
			\hline\hline
			& \multicolumn{3}{ c | |}{ approximate rank-2, $p = 100,  \upsilon = 20\% $} 
			& \multicolumn{3}{ c  }{ approximate rank-5, $p = 100,  \upsilon = 20\% $} 
			\\
			Errors   & Gibbs & LMC & MALA  
			& Gibbs & LMC & MALA 
			\\ \hline
			$10^1\times$MSE & 5.347 (.168) & 5.347 (.168) & 5.347 (.168) 
			& 5.933 (.128) & 5.933 (.128) &5.933 (.128) 
			\\ 
			$10^1\times$NMSE  & 2.254 (.244) & 2.255 (.244) & 2.254 (.244) 
			& 1.079 (.086) & 1.079 (.086) &1.079 (.086) 
			\\ 
			$10^1\times$Pred  & 5.803 (.266) & 5.804 (.266) & 5.804 (.266) 
			& 7.112 (.295) &  7.112 (.295) & 7.112 (.295) 
			\\
			\hline
			& \multicolumn{3}{ c ||}{ approximate rank-2, $p = 100,  \upsilon = 50\% $  } 
			& \multicolumn{3}{ c  }{approximate rank-5, $p = 100,  \upsilon = 50\% $} 
			\\  
			& LMC & MALA & Gibbs 
			& LMC & MALA & Gibbs
			\\ \hline
			$10^1\times$MSE & 5.936 (.188) & 5.936 (.188) & 5.936 (.188) 
			& 7.388 (.205) & 7.389 (.205) & 7.388 (.205)
			\\
			$10^1\times$NMSE	& 2.476 (.311) & 2.476 (.311) & 2.476 (.311) 
			& 1.320 (.107) & 1.321 (.107) & 1.321 (.107) 
			\\
			$10^1\times$Pred  & 6.424 (.253) & 6.424 (.253) & 6.424 (.253) 
			& 8.765 (.293) & 8.765 (.293) &8.766 (.293) 
			\\
			\hline
			& \multicolumn{3}{ c| |}{ approximate rank-2 $p = 100,  \upsilon = 80\% $  }
			& \multicolumn{3}{ c  }{ approximate rank-5 $p = 100,  \upsilon = 80\% $}  
			\\  
			& LMC & MALA & Gibbs 
			& LMC & MALA & Gibbs
			\\ \hline
			$10^1\times$MSE & 8.506 (.374) & 8.507 (.374) & 8.506 (.374) 
			& 19.529 (2.150) & 19.529 (2.150) & 19.529 (2.150) 
			\\
			$10^1\times$NMSE	& 3.466 (.404) & 3.467 (.404) & 3.466 (.404) 
			& 3.592 (.516) & 3.593 (.516) &3.593 (.516) 
			\\
			$10^1\times$Pred  & 9.095 (.431) & 9.096 (.431) & 9.095 (.431) 
			& 22.199 (2.472) & 22.200 (2.472) &22.200 (2.472) 
			\\ 
			\hline\hline
			& \multicolumn{3}{ c || }{approximate rank-2 $p = 500,  \upsilon = 20\% $} 
			& \multicolumn{3}{ c  }{approximate rank-5 $p = 500,  \upsilon = 20\% $} 
			\\
			& LMC & MALA & Gibbs 
			& LMC & MALA & Gibbs
			\\ \hline
			$10^1\times$MSE & 5.181 (.110) & 5.181 (.110) & 5.181 (.110) 
			& 5.446 (.124) & 5.446 (.124) & 5.446 (.124) 
			\\ 
			$10^1\times$NMSE  & 2.090 (.181) & 2.090 (.181) & 2.090 (.181) 
			& 0.998 (.076) & 0.998 (.076) &0.998 (.076) 
			\\ 
			$10^1\times$Pred & 5.460 (.142) & 5.460 (.142) & 5.460 (.142) 
			& 6.177 (.145) & 6.177 (.145) & 6.177 (.145) 
			\\
			\hline
			& \multicolumn{3}{ c ||}{approximate rank-2 $p = 500,  \upsilon = 50\% $  } 
			& \multicolumn{3}{ c  }{approximate rank-5 $p = 500,  \upsilon = 50\% $} 
			\\  
			& LMC & MALA & Gibbs 
			& LMC & MALA & Gibbs
			\\ \hline
			$10^1\times$MSE & 5.488 (.109) & 5.488 (.109) & 5.488 (.109) 
			& 6.339 (.146) & 6.339 (.146) & 6.339 (.146) 
			\\
			$10^1\times$NMSE	& 2.206 (.178) & 2.206 (.178) & 2.206 (.178) 
			& 1.167 (.074) &   1.168 (.074) & 1.167 (.074) 
			\\
			$10^1\times$Pred  & 5.762 (.128) & 5.762 (.128) & 5.762 (.128) 
			& 7.087 (.182) & 7.087 (.182) & 7.087 (.182) 
			\\
			\hline
			& \multicolumn{3}{ c| |}{approximate rank-2, $p = 500,  \upsilon = 80\% $  }
			& \multicolumn{3}{ c  }{ approximate rank-5, $p = 500,  \upsilon = 80\% $}  
			\\  
			& LMC & MALA & Gibbs 
			& LMC & MALA & Gibbs
			\\ \hline
			$10^1\times$MSE & 6.805 (.175) & 6.806 (.175) & 6.806 (.175) 
			& 10.648 (.240) & 10.648 (.240) &10.648 (.240) 
			\\
			$10^1\times$NMSE	& 2.776 (.278) & 2.776 (.278) &2.776 (.278) 
			& 1.957 (.139) & 1.957 (.139) &1.957 (.139) 
			\\
			$10^1\times$Pred  & 7.106 (.200) & 7.107 (.200) & 7.106 (.200) 
			& 11.666 (.285) &  11.667 (.285) & 11.666 (.285) 
			\\ \hline\hline
		\end{tabular}
		\label{tb_setting2}
	\end{table}

	\begin{table}[!h]
		\footnotesize
		\caption{Results for image inpainting with data Lena. The mean and the standard deviation (in parentheses) of each error between the simulation replicates are presented. (MSE: average of mean square errors; Rank: average of estimated ranks; NMSE: average of normalized mean square errors; Pred: average of prediction errors.)}
		\begin{tabular}{p{17mm}|ccc||ccc} 
			\hline\hline
			& \multicolumn{3}{ c | |}{$\upsilon = 50\% $} 
			& \multicolumn{3}{ c  }{$ \upsilon = 20\% $} 
			\\
			Errors   & Gibbs & LMC & MALA  
			& Gibbs & LMC & MALA 
			\\ \hline
			MSE & 382.51 (1.37) & 136.91 (2.32) & 136.91 (2.32) 
			& 350.91 (0.33) & 28.23 (.71) & 28.23 (.71)
			\\ 
			Rank  & 3.93 (0.25) & 3 (0)  &  3 (0) 
			& 3.5 (0.5) & 3 (0)  &  3 (0) 
			\\ 
			NMSE  & .0216 (.0001) & .0077 (.0001)  & .0077 (.0001) 
			& .01989 ($2.10^{-5}$) & .00160 ($4.10^{-5}$) & .00160 ($4.10^{-5}$)
			\\ 
			Pred  & 453.95 (4.09) & 273.10 (4.56)  &  273.10 (4.56)
			& 415.38 (7.41) & 139.49 (3.28) & 139.49 (3.28) 
			\\ \hline\hline
		\end{tabular}
		\label{tb_lena}
	\end{table}

	\clearpage
	\appendix
	
	\section{Appendix}
	\label{appendix}
	\label{sc_proofs}
	\subsection{Brief review on futher related works}
	It is worth noting that most studies in matrix completion consider that the sampling distribution is uniform, see \cite{alquier2014bayesian, candes2010matrix, candes2009exact,
		candes2010power, koltchinskii2011nuclear,alquier2020concentration,lim2007variational} among others.  However, in practice the observed entries are not assured to be uniformly distributed: for example, some items are more popular  and as a consequence receive much more ratings than others. More importantly,  it is noted that the sampling distribution is usually not  known in practice.  Several studies have been performed for general sampling schemes rather than uniform distribution, see  e.g. \cite{foygel2011learning, klopp2014noisy,negahban2012restricted}.  Moreover, the paper \cite{mai2015} is the first line of works that studied (Bayesian) matrix completion under a general sampling model without any restriction.  
	
	The spectral scaled Student prior has been considered before in the context of image denoising  \cite{dalalyan2020exponential} and in reduced rank regression \cite{mai2021efficient}.  Moreover,  it is shown that if a random matrix $ M $ has the density distribution as in \eqref{prior_scaled_Student}, then the random vectors $ M_i $ are all drawn from the $ p$-variate scaled Student distribtuion $ (\tau /\sqrt{3})t_{3,p} $, \cite[Lemma 1]{dalalyan2020exponential}. On the other hand,  this prior can be seen as the marginal distribution of the Gaussian-inverse Wishart prior that is explored in \cite{yang2018fast} in the context of matrix completion where the precision matrix is integrated out. Therefore, in some ways, our work provide theoretical assessment for that paper. However, we note that the paper \cite{yang2018fast} consider an opimization approach to Bayesian matrix completion using variational inference while our work propose an efficient sampling approach.

	\subsection{Short review on low-rank factorization priors}
	Up to our knowledge, the paper \cite{geweke1996bayesian} was the first work that introduced a low-rank factorization prior in a context of low-rank matrix estimation. The idea is to express the matrix $M $ as $ M = U_{m\times k} V_{p\times k}^\top$ with $k \leq \min(p,m) $ and then the prior is difined on $U $ and $V $ rather than on $M $ as
	$$
	\pi(M,N) \propto 
	\exp\left\{ - \frac{\tau^2}{2} ( \|M\|^2_F + \|N\|^2_F)   \right\} 
	$$
	for some $\tau>0$. This approach has been used in matrix completion for the first time in \cite{lim2007variational}.  However,  this approach is faced with the problem of choosing $k$.  One could perform model selection for any possible $k$ as done in \cite{ kleibergen2002priors,corander2004bayesian} but the computation may be expensive for large data set.  Recent approaches as in \cite{babacan2012sparse} focus on fixing a large $k$, e.g. $k = \min (p,m)$, then sparsity-promoting priors are placed on the columns of $U $ and $V $ such that most columns are almost null.  As a consequence,  the resulting matrix $ M = UV^\top$ is approximately low-rank.  See \cite{alquier2013bayesian,alquier2014bayesian} for the details and dicussions on low-rank factorization priors.  
	
	Using low-rank factorization priors, most authors use the Gibbs sampler to simulate from the posterior as the conditional posterior distributions can be  derived explitcitely, e.g. in \cite{salakhutdinov2008bayesian}. However, it is noted that these Gibbs sampling algorithms update the factor matrices in a row-by-row manner and invole a number of matrix inverse or singular value decomposition operations at each iteration.  This is expensive and slow down the algorithm for large data set significantly.

	\subsection{Theoretical result in special cases: uniformly sampling and exact low-rank}
	
	Firstly, when the sampling distribution is uniform, we obtain the following result for the Frobenius norm directly from Theorem \ref{thrm_main}.
	
	\begin{corollary}
		Let Assumption \ref{bounded assume} and \ref{bruit-Pac} be satisfied and take $ \lambda = \lambda^* := n/C $. Then, with probability at least $ 1 - \epsilon, \epsilon\in (0,1) $, one has for any matrix $ \bar{M} $ with its rank at most $r$ that
		\begin{equation*}
		\dfrac{  \| \hat{M}_{\lambda^*} - M^*  \|^2_{F}  }{mp}
		\leq
		\inf_{ \bar{M} } \,
		(1 + \delta)  \dfrac{ \| \bar{M} - M^*  \|^2_{F} }{mp}
		+ 
		\frac{ \mathcal{C}  r(m+p) \log \left(mp \right)  } {n}
		+
		\frac{\mathcal{C} } {n}\log\frac{2}{\varepsilon}
		,
		\end{equation*}
		where $ \delta $ is in $ (0,1) $ and $ \mathcal{C} >C $  is a universal constants that depends on $ \sigma^2, L, \xi $ only.
	\end{corollary}
	
	Secondly,  as soon as the true matrix $ M^* $ is low-rank where its rank is atmost $ r $,  we can pick $ \bar{M} = M^* $ and obtain the following results.
	
	\begin{corollary}
		Let Assumption \ref{bounded assume} and \ref{bruit-Pac} be satisfied and take $ \lambda = \lambda^* := n/C $. Then, with probability at least $ 1 - \epsilon, \epsilon\in (0,1) $, one has 
		\begin{equation}
		\| \hat{M}_{\lambda^*} - M^*  \|^2_{F,\Pi} 
		\leq
		\frac{ \mathcal{C}  r(m+p) \log \left(mp \right)  } {n}  
		+
		\frac{\mathcal{C} } {n}\log\frac{2}{\varepsilon}
		,
		\end{equation}
		where $ \delta $ is in $ (0,1) $ and $ \mathcal{C} >C $  is a universal constants that depends on $ \sigma^2, L, \xi $ only.
	\end{corollary}

	\subsection{A Metropolis-adjusted Langevin algorithm}

	Here, we propose a Metropolis-Hasting correction to the LMC. This approach guarantees the convergence to the posterior and it also provides a useful way for choosing $ h $. More precisely, we consider the update rule in \eqref{langevinMC} as a proposal for a new state,
	\begin{align}
	\tilde{M}_{k+1} = M_{k} - h\nabla \log    \hat{\rho}_{\lambda}  (M_k) +\sqrt{2h}\,W_k,\qquad
	k=0,1,\ldots,
	\label{mala}
	\end{align}
	Note that $\tilde{M}_{k+1} $ is normally distributed with mean $ M_{k} - h\nabla \log    \hat{\rho}_{\lambda}  (M_k) $ and the covariance matrices equal to $ 2h $ times the identity matrices. This proposal is accepted or rejected according to the Metropolis-Hastings algorithm that the proposal is accepted with probabiliy:
	$$
	\min \left\lbrace 1,  \frac{  \hat{\rho}_{\lambda}  (\tilde{M}_{k+1}) q(M_k | \tilde{M}_{k+1}) }{
		\hat{\rho}_{\lambda}  (M_k ) q(\tilde{M}_{k+1} | M_k ) } \right\rbrace,
	$$
	where 
	$$
	q(x' | x) \propto \exp \left(-\frac{1}{4h}\|x'-x + h\nabla \log    \hat{\rho}_{\lambda}  (x) \|^2_F \right)
	$$
	is the transition probability density from $x$ to $x'$.  Compared to random-walk Metropolis–Hastings,  the advantage of MALA is that it usually proposes moves into regions of higher probability, which are then more likely to be accepted.

	We note that the step-size $h$ is chosen such that the acceptance rate is approximate $0.5$ following \cite{roberts1998optimal}. See Section \ref{sc_numer} for some choices in special cases in our simulations.

	It is also noted that an immediate application of the LMC algorithm in \eqref{langevinMC} needs to calculate a $p\times p$ matrix inversion at each iteration.  This might be expensive and can slow down significantly the algorithm.  Therefore, in the case of very big data or huge $p $, we could replace this matrix inversion by its accurately approximation through a convex optimization.  More precisely,  it can be easily verified that the matrix
	$\mathbf{B} = (\tau^2 \mathbf{I}_{m} + MM^\intercal )^{-1} M $ is the solution to the following convex optimization problem
	$$
	\min_{\mathbf{B} } \big\{\| \mathbf{I}_p- M^\top \mathbf{B}  \|_F^2 + \tau^2\|\mathbf{B} \|_F^2\big\}.
	$$
	The solution of this optimization problem can be conveniently obtained by using the package 'glmnet' \cite{glmnet} (with family option 'mgaussian'). This avoid to perform matrix inversion or other costly calculation.  However,  we note here that the LMC algorithm is being used with approximate gradient evaluation, theoretical assessment of this approach can be found in \cite{dalalyan2019user}.

	\subsection{Proofs}
	
	The main argument of the proof is based on the so-called PAC-Bayesian bounds which were introduced in \cite{STW,McA}.  However,  our results are derived by following the Catoni’s works \cite{MR2483528} where the author shown how to derive powerful oracle inequalities from PAC-Bayesian bounds.  Since then this technique has been use many times to obtain oracle inequalities in many different problems \cite{dalalyan2008aggregation,
		mai2015,
		alquier2011pac,
		guedj2018pac,
		ridgway2014pac,
		cottet20181,
		mai2017pseudo,
		alquier2016properties}
	
	For the sake of simplicity, let us define
	\begin{align*}
	\alpha = \left(\lambda
	-\frac{\lambda^{2} C_1  }{2n(1-\frac{ C_2 \lambda}{n})}\right) ;
	\quad
	\beta = \left(\lambda
	+\frac{\lambda^{2} C_1 }{2n(1-\frac{ C_2 \lambda}{n})}\right) .
	\end{align*}
	In order to understand what follows, keep in mind that  our optimal estimator comes with $ \lambda = \lambda^* = \frac{n}{C } $, so $ \alpha $ and $ \beta $ are of order $ n $.
	
	First, we state  a general PAC-Bayesian inequality for matrix completion from \cite{mai2015},  in the style of~\cite{catoni2004statistical,dalalyan2008aggregation}.
	\begin{theorem}
		\label{thrm_PACBayes}
		Let Assumption \ref{bounded assume} and \ref{bruit-Pac} be satisfied. Then, for any $\epsilon\in (0,1)$, with probability at least $1 - \epsilon$, one has
		\begin{equation}
		\label{PAC_Bayes_bound}
		\| \hat{M}_{\lambda} - M^*  \|^2_{F,\Pi} 
		\leq
		\inf_{\rho\in\mathfrak{M}_{+}^{1}(M) } \frac{ 
			\beta  \int  \| M - M^*  \|^2_{F,\Pi} d\rho
			+ 
			2 [KL(\rho, \pi) + \log\frac{2}{\varepsilon} ] } {
			\alpha},
		\end{equation}
		where $\mathfrak{M}_{+}^{1}(M)$ is the set of all positive
		probability measures over $M$.
	\end{theorem}
	This theorem is the result from the Step 1 in the proof of Theorem 1 in \cite{mai2015}. 
	
	Now, we are ready to present the proof of Theorem \ref{thrm_main}.
	
	\begin{proof}[\textbf{ of Theorem \ref{thrm_main}}]
		Let's fix an arbitrary matrix $ \bar{M} $.  Let $ \bar{\rho} $ be the distribution obtained from the prior $ \pi $ by a translation
		$$
		\bar{\rho} (M) = \pi ( M - \bar{M} ) .
		$$
		Now, we apply Theorem \ref{thrm_PACBayes} and upper bound the right hand side of \eqref{PAC_Bayes_bound} evaluated at the distribution $ \bar{\rho} $. We obtain
		\begin{align*}
		\| \hat{M}_{\lambda} - M^*  \|^2_{F,\Pi} 
		\leq
		\frac{ \beta}{\alpha}  \int  \| M - M^*  \|^2_{F,\Pi} d \bar{\rho}
		+ 
		\frac{2 } {\alpha}  \left[ KL( \bar{\rho} ,  \pi) + \log\frac{2}{\varepsilon} \right]  .
		\end{align*}
		Firstly, we deal with the integral term that
		\begin{align*}
		\int  \| M - M^*  \|^2_{F,\Pi} d \bar{\rho}
		=
		\int  \| M - M^*  \|^2_{F,\Pi}  \pi ( M - \bar{M} ) dM.
		\end{align*}
		By using the translation invariance of the Lebesgue measure and the fact that $ \int M \pi(M)dM = 0 $, we obtain
		\begin{align*}
		\int  \| M - M^*  \|^2_{F,\Pi}  \pi ( M - \bar{M} ) dM
		=
		\| \bar{M} - M^*  \|^2_{F,\Pi} 
		+
		\int  \| M \|^2_{F,\Pi}  \pi ( M ) dM .
		\end{align*}
		For the last integral, we have that
		\begin{align*}
		\int  \| M \|^2_{F,\Pi}  \pi ( M ) dM 
		\leq 
		\int \sum_{i=1}^m \| M_i \|^2_{F}  \pi ( M ) dM 
		=
		mp \tau^2 ,
		\end{align*}
		using the Lemma 1 in \cite{dalalyan2020exponential}.
		
		Secondly, we deal with the Kullback-Leibler term. We have that 
		\begin{align*}
		KL( \bar{\rho} ,  \pi) 
		\leq  &
		2 {\rm rank} (\bar{M}) (m+p+2) \log \left( 1+ \frac{\| \bar{ M } \|_F}{\tau \sqrt{2{\rm rank} (\bar{M})}} \right) 
		\\
		\leq  &
		2 {\rm rank} (\bar{M}) (m+p+2) \log \left( 1+ \frac{ \sqrt{mp} }{\tau \sqrt{2{\rm rank} (\bar{M})}} \right) 
		\end{align*}
		that is followed by using Lemma 2 in  \cite{dalalyan2020exponential}.
		
		Therefore, we have for any matrix $ \bar{M} $ with $ {\rm rank}( \bar{M} ) = r $ that
		\begin{align}
		\label{bounds}
		\| \hat{M}_{\lambda} - M^*  \|^2_{F,\Pi} 
		\leq
		\frac{ \beta}{\alpha} &\left( \| \bar{M} - M^*  \|^2_{F,\Pi}  
		+ mp \tau^2 \right)
		+ \nonumber
		\\
		&
		\frac{4 } {\alpha}  r(m+p+2) \log \left( 1+ \frac{ \sqrt{mp} }{\tau \sqrt{2 r  }} \right)
		+
		\frac{2}{\alpha}  \log\frac{2}{\varepsilon}
		.
		\end{align}
		
		Taking now $ \lambda = \lambda^* := n/C $ with $ C > C_2 + 3C_1/2 $,  we get that
		$
		\frac{\beta}{\alpha} \leq 1 + \delta
		$
		with $  0<\delta < 1 $, and that $   \frac{4 } {\alpha} \leq 6C / n $. Thus, the inequality \eqref{bounds} yeilds
		\begin{align*}
		\| \hat{M}_{\lambda} - M^*  \|^2_{F,\Pi} 
		\leq
		&
		(1 + \delta) \| \bar{M} - M^*  \|^2_{F,\Pi} 
		+ 2mp \tau^2
		\\
		& + 
		\frac{6C } {n}  r(m+p+2) \log \left( 1+ \frac{ \sqrt{mp} }{\tau \sqrt{2 r  }} \right)
		+
		\frac{3C } {n}\log\frac{2}{\varepsilon}
		.
		\end{align*}
		Now, by taking $ \tau^2 = r(m+p)/(mpn) $, we obtain that 
		\begin{align*}
		\| \hat{M}_{\lambda} - M^*  \|^2_{F,\Pi} 
		\leq
		(1 + \delta) \| \bar{M} - M^*  \|^2_{F,\Pi} 
		+ 
		\frac{ \mathcal{C} } {n}  r(m+p) \log \left(mp \right)
		+
		\frac{\mathcal{C} } {n}\log\frac{2}{\varepsilon}
		,
		\end{align*}
		where $  \mathcal{C} > 6C $ is a universal constants that depends on $ \sigma^2, L, \xi $ only.
		
		This completes the proof of the theorem.
	\end{proof}

\end{document}